%% file: main.tex
\documentclass[10pt,twocolumn,letterpaper]{article}

\usepackage[pagenumbers]{cvpr} 
\definecolor{cvprblue}{rgb}{0.21,0.49,0.74}
\usepackage[pagebackref,breaklinks,colorlinks,allcolors=cvprblue]{hyperref}
\usepackage{siunitx}
\usepackage{booktabs}
\usepackage{multirow}
\usepackage{float}
\usepackage{cuted}
\usepackage{booktabs,siunitx}
\usepackage{fontawesome5}
\usepackage[table]{xcolor}
\usepackage{comment}

\def\confName{CVPR}
\def\confYear{2026}

\title{CoFiDA-M: Concept-Aware Feature Modulation for Cross-Domain Adaptation with Image-Only Inference}

\author{Nurjahan Sultana, Moi Hoon Yap, Xinqi Fan, Wenqi Lu\textsuperscript{\dag}\\
Department of Computing and Mathematics, Manchester Metropolitan University\\
Dalton Building, Chester Street, M1 5GD Manchester, UK\\
{\tt\small nurjahan.sultana@stu.mmu.ac.uk; \{m.yap, x.fan, w.lu\}@mmu.ac.uk}
}

\newenvironment{tightmath}{%
  \begingroup
  \setlength{\abovedisplayskip}{2pt}%
  \setlength{\belowdisplayskip}{2pt}%
  \setlength{\abovedisplayshortskip}{1pt}%
  \setlength{\belowdisplayshortskip}{1pt}%
  \setlength{\jot}{2pt}% for align row spacing
  \setlength{\parskip}{0pt}% stop extra para gap around equations
}{\endgroup}

\begin{document}
\maketitle
\begingroup
\renewcommand{\thefootnote}{\dag}
\footnotetext{Corresponding author: w.lu@mmu.ac.uk}
\endgroup

\begingroup
\renewcommand{\thefootnote}{}
\footnotetext{This is the arXiv version of a paper accepted to CVPR 2026.}
\endgroup

\begin{abstract}
Models for AI-based skin cancer screening suffer a severe performance drop when shifting from expert dermoscopic (source) images to consumer-grade clinical (target) images, hindering real-world deployment. Existing domain adaptation methods often ignore crucial semantic invariants, such as clinical concepts. While new foundation models like MONET can provide this semantic information as dense, probabilistic scores, this metadata is unavailable at test time, creating a deployment paradox for practical image-only screening tools. We address this gap by proposing CoFiDA-M, a privileged information framework that learns from concepts at training time but deploys as an image-only model. Our method trains a teacher network that uses MONET concept probabilities to guide a FiLM modulator, transforming visual features into a semantically ``edited" feature space. A lightweight, image-only student is then trained to reproduce this edited representation, not just the teacher's final predictions. This distillation ``bakes" the clinical reasoning into the student's weights.
On a challenging multi-dataset benchmark, our image-only student significantly outperforms state-of-the-art approaches, especially in melanoma recall. Our work provides a practical and generalizable framework for leveraging noisy, probabilistic metadata as privileged information, demonstrating strong cross-dataset robustness and potential for real-world deployment beyond dermatology. Implementation code is available at \href{https://github.com/mmu-dermatology-research/CoFiDA.git}{GitHub}
.
\end{abstract}
\vspace{-0.8em}

\begin{figure}[t]
\captionsetup{aboveskip=2pt, belowskip=-15pt}
    \centering    \includegraphics[width=0.47\textwidth,height=0.15\textheight]{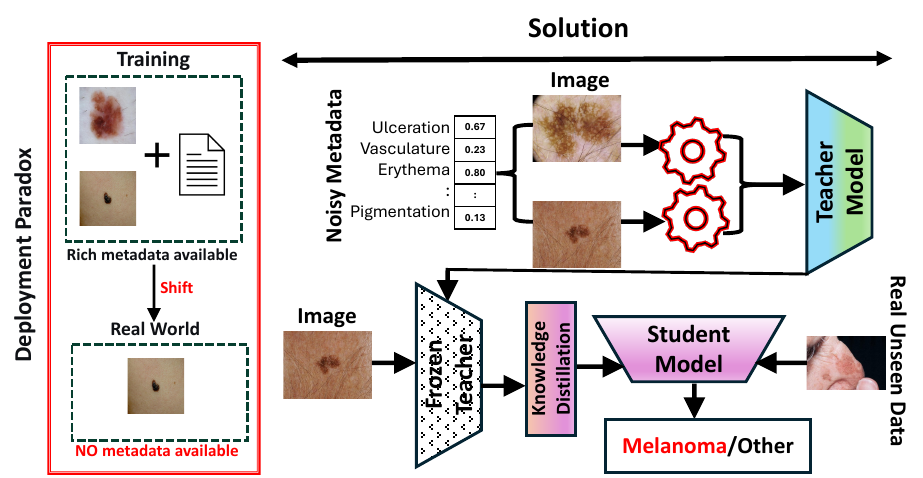}
    \caption{\textbf{CoFiDA-M} resolves the \emph{Deployment Paradox}: metadata are available at training but absent in the real world. A concept-guided teacher trained with noisy metadata transfers its reasoning to an image-only student that generalizes to unseen clinical data.}
    \label{theme}
\end{figure}
%\vspace{-13pt}
\section{Introduction}
%\vspace{-3pt}
The promise of AI for accessible skin cancer screening on mobile phones is hindered by a critical domain shift, as models trained on expert dermoscopic images degrade sharply on clinical smartphone photos \cite{chamarthi2024mitigating, wang2024achieving, sultana2025selective}. This is not a static gap that can be solved by simply adding more clinical images to the training set. Every new camera, lighting condition, and patient population effectively behaves as a novel domain, causing models to fail when encountering these unseen variations \cite{ovadia2019can,sultana2025domain}. For AI to be reliably deployed in the wild, it must learn domain invariant features ensuring robust performance across all target distributions.

To bridge this domain gap, Unsupervised Domain Adaptation (UDA) methods aim to adapt a model from a labeled source domain to an unlabeled target domain. Many classical approaches achieve this by forcing the source and target feature distributions to become indistinguishable, often by using adversarial training \cite{ganin2016domain} or directly matching statistical moments like covariance \cite{sun2016deep} and the maximum mean discrepancy \cite{gretton2012kernel} across multiple layers. The shared goal of these methods is to learn a single, unified feature space for both source and target data. 

However, in high stakes medical domains, this global alignment is often insufficient. It ignores crucial semantic invariants, high level clinical concepts that are shared across domains even when the low level pixel data differs. A clinical concept like `ulceration' or a `pigment network' remains the same, whether viewed in a dermoscopic image or a standard clinical photo. A model that aligns domains based on these shared medical concepts would be more robust and interpretable than one that simply matches global statistics. This aligns with findings in multimodal medical AI, where grounding models in clinical knowledge and concepts, as seen in MedCLIP \cite{wang2022medclip} or PanDerm \cite{yan2025multimodal}, has proven critical for building robust systems. 

Recent research has increasingly utilized language and high-level conceptual information to inform and guide the adaptation process, as exemplified by the PØDA framework \cite{fahes2023poda}, which shows that a simple text prompt (``driving at night") can steer features to adapt to a new, unseen domain in a zero-shot fashion. LAGUNA \cite{diko2024laguna} takes this further by using language to align the relative geometry of the class structure between domains, rather than forcing an absolute feature overlap. While powerful, these methods have limitations, i.e., PØDA is designed for text prompted zero shot transfer, and LAGUNA critically relies on having text captions available for the target domain during adaptation \cite{diko2024laguna}. In dermatology, concept-based models such as MAKE \cite{yan2025make} inject structured expert clinical knowledge during pretraining to enable zero shot lesion recognition, but they do not solve unsupervised domain adaptation between dermoscopic and clinical images. This gap is particularly acute in specialized fields like medicine, where recent tools like MONET \cite{kim2024transparent}, a foundation model grounded in medical literature, can now provide dense, probabilistic scores for high level concepts (e.g., `ulceration', `pigmentation'). Such fine grained, probabilistic metadata offers a much richer supervisory signal than simple text prompts, but has been largely overlooked by domain adaptation research, likely due to its prior unavailability. The availability of such rich metadata, however, gives rise to a deployment paradox. Although these detailed annotations can enhance model performance during training, the ultimate goal of accessible mobile screening necessitates a simple, image-only model, since patients cannot supply expert-level metadata at inference time. Consequently, a key practical challenge remains. While this problem falls under the Privileged Information (PI) setting recently formalized for UDA by DALUPI \cite{Breitholtz2024DALUPI}, its proposed two-stage ``hallucination" pipeline is not designed for our specific challenge. How to apply PI effectively in our medical context is not fully addressed. The key challenges are: (1) the concepts are available only during training, (2) the PI is a ``messy" and noisy probability vector, not clean text or attributes, and (3) the final model must be a lightweight, image-only model that works on an unlabeled target domain.

To address this specific gap, we introduce CoFiDA-M: Confidence Gated Feature Editing for Domain Adaptation. We propose a teacher-student framework that leverages external concept probabilities (like MONET scores) as privileged information during training, while ensuring the final student model is image-only at inference. Figure \ref{theme} illustrates the deployment paradox and a high-level overview of our proposed solution. Our key contributions are:

\begin{itemize}
    \item We introduce CoFiDA-M, a novel framework that integrates MONET-derived dermatological concepts into a FiLM based modulation mechanism, aligning visual features with semantic clinical context during training.
    \item We propose a novel distillation scheme where the student learns to match not only the teacher's final predictions (logits) but also its edited feature, allowing the student to learn the reasoning process, not just the final prediction.
    \item Unlike previous concept based models that require concept inputs at test time, CoFiDA-M transfers the semantic alignment into a concept-independent student. At inference, only image inputs are required, enabling faster and more practical deployment while maintaining both interpretability and accuracy without concept supervision.
    \item Our model demonstrates superior generalization and interpretability: validated on a challenging out-of-domain benchmark of six unseen datasets, our image-only student outperforms state-of-the-art methods with significant improvements in AUROC and melanoma recall, while preserving interpretability through the learned concept feature modulation dynamics.
\end{itemize}

%\vspace{-15pt}
\section{Related Work}
%\vspace{-3pt}
Recent multimodal domain adaptation leverages language and expert concepts to steer visual representations across shifts. Vision language pre-training (VLP) provides a shared image text space that can be adapted with lightweight prompts, enabling transfer without full target supervision. In medical imaging, this idea appears in general VLMs \cite{li2022blip}, radiology-specific models like MedCLIP \cite{wang2022medclip}, and comprehensive multimodal reasoning systems like MIMO \cite{chen2025mimo}. Dermatology specific resources further strengthen this route by pairing images with rich text and structured clinical descriptors, as seen in the PanDerm foundation model \cite{yan2025multimodal} and the MM-Skin/SkinVL dataset \cite{zeng2025mm}, improving the zero shot baselines that many Domain Adaptation (DA) pipelines build upon. Standard classification models also show that combining skin lesion images with clinical metadata improves multimodal learning \cite{zhang2023tformer,christopoulos2025skin}.

Building on the use of language, prompt-driven adaptation uses textual cues to shift decision boundaries, ``steering" features towards a new domain without target labels. This strategy shows that a global text prompt can deliver zero shot gains under appearance change \cite{fahes2023poda}. More advanced methods use language to align the relative geometry of class structures \cite{diko2024laguna} or learn domain-agnostic representations through mutual text visual prompting \cite{du2024domain}. While powerful, these approaches typically require clean text prompts or target domain captions, and do not provide per-image, calibrated evidence to supervise finer, concept level edits.

A complementary line of research conditions the visual pathway directly on concepts or attributes. This approach aligns with the learning using PI paradigm, recently formalized for UDA as DALUPI \cite{Breitholtz2024DALUPI}. This framework leverages auxiliary data (such as concepts) that is available only during training to improve adaptation \cite{Breitholtz2024DALUPI}. However, the method proposed by DALUPI relies on a two-stage pipeline, where one model learns to ``hallucinate'' the PI ($X \rightarrow W$), and a second model predicts from that hallucination ($W \rightarrow Y$) \cite{Breitholtz2024DALUPI}. In contrast, we propose a more direct and unified strategy. Our CoFiDA-M framework uses the privileged concepts to directly modulate the visual feature space via a FiLM-based teacher network \cite{perez2018film}. Critically, we introduce a novel distillation scheme where an image-only student learns to reproduce the teacher's entire semantically-edited feature representation, rather than just mimicking its final predictions. This conditioning logic is also actively used in contemporary medical VLP, such as in the knowledge enhanced modules of models like MAKE for zero shot dermatology assessment \cite{yan2025make}.

Beyond multimodal approaches, other UDA methods address alternative data constraints. A prominent direction focuses on Source-Free Domain Adaptation (SFDA), where a pre-trained source model is adapted using only unlabeled target data, a key constraint in medical applications. These approaches often rely on generating pseudo-labels by discovering target class prototypes, as in CPD \cite{zhou2024source}, or by freezing the source classifier and training the feature encoder with information maximization, a method known as ``hypothesis transfer" or SHOT \cite{liang2020we}. Other SFDA methods adapt from black box APIs by distilling predictions, like DINE \cite{liang2022dine}, or, for segmentation, synthesize source like images by matching the source model’s BatchNorm statistics to enable knowledge transfer (SFDA) \cite{liu2021source}. Concurrently, classical UDA methods (which retain access to source data) also continue to evolve, with IT-RUDA \cite{rashidi2025ruda} aligning source and target features using a robust $\alpha$-divergence that down-weights outliers and stabilizes transfer under label or sample imbalance \cite{rashidi2025ruda}.

A related setting is Test-Time Adaptation (TTA), where the model adapts during inference as new target data arrives. A widely used consistency framework is the mean-teacher model, which forms a teacher by exponential moving average of the student’s weights and trains the student to match the teacher’s predictions on unlabeled data \cite{tarvainen2017mean}. The foundational TTA method, TENT \cite{wang2021tent}, minimizes prediction entropy by only updating normalization layer parameters. Building on this, CoTTA \cite{wang2022continual} addresses adaptation in continually changing environments by using a mean teacher for stable pseudo-labels and stochastically restoring weights to prevent catastrophic forgetting. Complementary to these, the Wisdom-of-Crowds filter retains only samples whose confidence increases relative to the frozen source model, improving robustness under open set and streaming conditions \cite{lee2023towards}.

Our work bridges two strands of multimodal domain adaptation: prompt or language guided transfer \cite{li2022blip,wang2022medclip,fahes2023poda,diko2024laguna} and concept conditioned feature editing via lightweight modulation \cite{perez2018film,yan2025make}. We train with calibrated clinical concept probabilities to supervise edits in the multimodal teacher and then distil this behavior into an efficient, image-only student for deployment, framing the hand off as knowledge distillation \cite{hinton2015distilling}. This directly addresses the common gap between multimodal training signals and unimodal test-time constraints.

{\sloppy
\section{Methodology}
\label{sec:method}
\begin{tightmath}
Our proposed CoFiDA-M uses concept probabilities from MONET \cite{kim2024transparent} as PI during training to guide feature editing. The model learns how such concepts reshape the internal representation in a concept-guided teacher, and transfers this ability to an image-only student through distillation, enabling robust generalization without test-time adaptation. As described in Figure \ref{CoFIDA-M-ARCH}, the proposed framework consists of three stages:
\begin{itemize}
\item Teacher training: a MONET conditioned teacher edits image features via FiLM modulation and is trained with both labeled source and unlabeled target data.
\item Knowledge distillation: a student model learns to mimic the teacher’s concept-aware predictions and representations using only image inputs.
\item Inference: the distilled student model performs image-only classification.
\end{itemize}

\begin{figure*}[h]
\captionsetup{aboveskip=2pt, belowskip=-15pt}
    \centering
\includegraphics[width=1\textwidth]{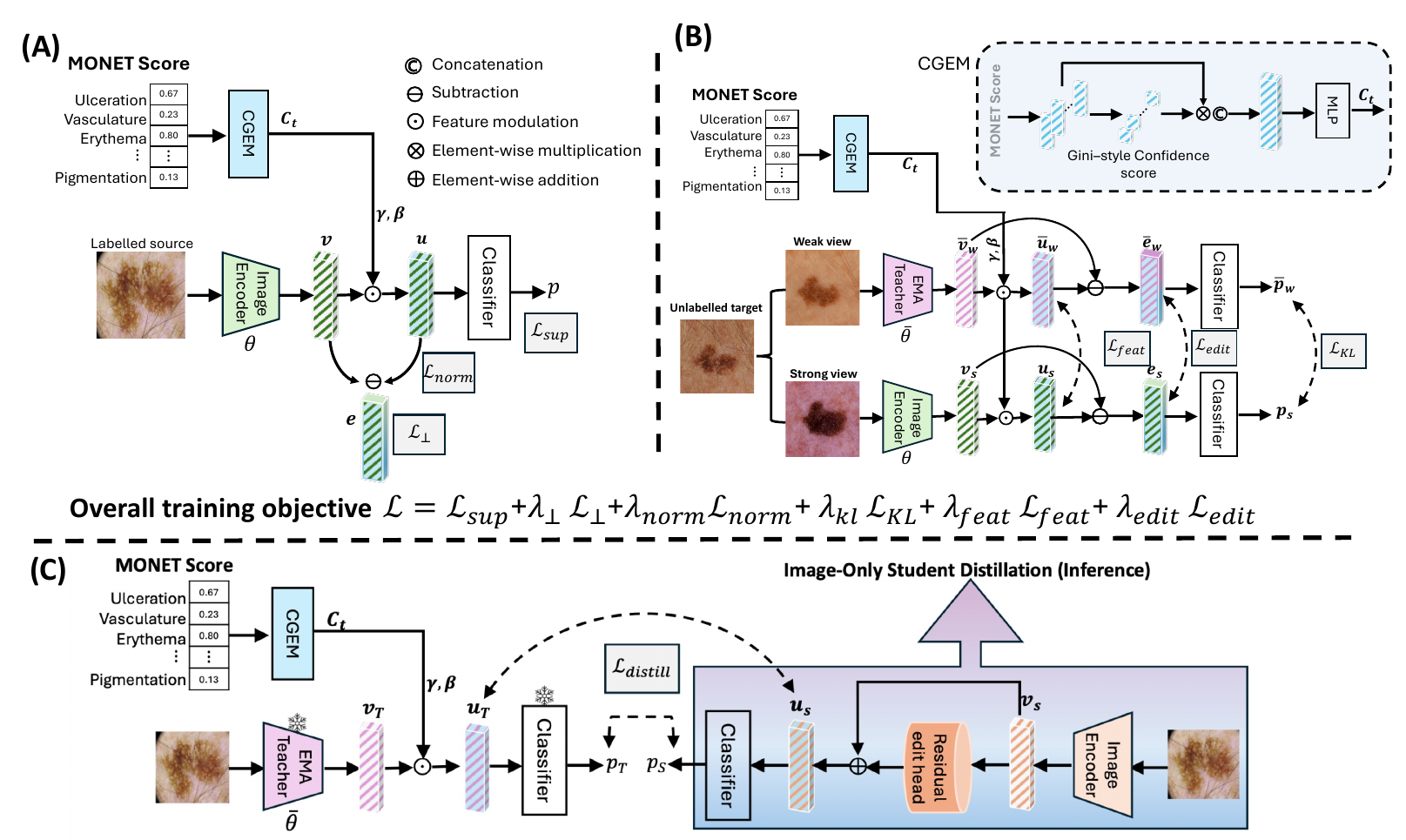}
    \caption{\textbf{Overview of the CoFiDA-M Framework.} Our method has three parts. (A) \& (B) Stage 1: Teacher Training. A concept-guided Teacher model is trained using labeled source data (A) and unlabeled target data (B). The target branch uses a Mean Teacher framework to align domains via multi-level consistency. (C) Stage 2: Distillation \& Stage 3: Inference. The ``expert'' Teacher is frozen (\faSnowflake{}) and its knowledge is distilled ($\mathcal{L}_{\text{distill}}$) into a lightweight Image-Only Student, which replaces the FiLM module with a residual edit head and is used for final inference.}
  \label{fig:cofida_framework}
    \label{CoFIDA-M-ARCH}
\end{figure*}
\subsection{Problem Setup}
We consider a labeled \emph{source} domain
$\mathcal{D}_s=\{(x_s^{(i)},\,s_s^{(i)},\,y_s^{(i)})\}_{i=1}^{N_s}$
consisting of dermoscopic images $x_s^{(i)}\!\in\!\mathbb{R}^{3\times H\times W}$ together with MONET concept probabilities and binary melanoma labels. The \emph{target} domain
$\mathcal{D}_t=\{(x_t^{(j)},\,s_t^{(j)})\}_{j=1}^{N_t}$ contains clinical images with associated MONET probabilities but no labels. An EfficientNet-B2 backbone \cite{tan2019efficientnet} $f_\theta$ embeds each image into a $d=1408$-dimensional penultimate feature space, $\mathbf{v}=f_\theta(x)$. Our goal is to exploit the labeled source data and the unlabeled target data to train a model that can predict the melanoma label $y$ for target images $x_t$, while operating with image-only inputs at inference.

%***********************

\subsection{Confidence-Gated MONET Embedding}
\label{sec:concept}
Each MONET attribute $a$ provides a probability $s_a\in[0,1]$ representing the likelihood of concept presence. Instead of using $s_a$ directly, we form a two–state probability vector 
\begin{equation}
\mathbf{q}_a=\begin{bmatrix}1-s_a\\ s_a\end{bmatrix}\in\mathbb{R}^2,
\end{equation}
which retains information about both concept presence and absence. To reflect uncertainty, we introduce a confidence gate embedding module (CGEM) on Gini impurity \cite{breiman2017classification}
\begin{equation}
g_a=s_a^2+(1-s_a)^2,\ \ 
\end{equation}
so that uncertain concepts ($s_a \approx 0.5$) contribute less. Each concept has a learnable embedding table $\mathbf{E}_a\!\in\!\mathbb{R}^{2\times 32}$, which is gated and projected using its query $\mathbf{q}_a$ and gate $g_a$. Concatenating these gated concept embeddings and passing them through a two–layer MLP yields the clinical conditioning vector
\begin{equation}
\mathbf{c}_t=\mathrm{MLP}\!\left(\operatorname{concat}_a\{\,g_a(\mathbf{E}_a^{\top}\mathbf{q}_a)\,\}\right)\in\mathbb{R}^{256},
\end{equation}
which summarizes the high-level clinical semantics.
\vspace{-0.4cm}
\subsection{Concept$-$Guided Feature Editing (Teacher)}
\label{sec:edit}
Within the teacher, clinical context from MONET guides the feature-editing operation preceding classification. The context vector $\mathbf{c}_t$ is used to modulate the image feature $\mathbf{v}$ through a Feature wise Linear Modulation (FiLM)\cite{perez2018film} based editing network. Specifically, a conditioning function $\boldsymbol{\psi}:\mathbb{R}^{256}\!\rightarrow\!\mathbb{R}^{2d}$ is used to predict clinical feature conditioned affine parameters 

\begin{equation}
\boldsymbol{\gamma},\,\boldsymbol{\beta}
= \mathrm{split}\!\big(\boldsymbol{\psi}(\mathbf{c}_t)\big) 
\end{equation}
which are split into two $d$-dimensional vectors controlling scaling and shifting of $\mathbf{v}$. With these two affine parameters, the clinical edited image feature $\mathbf{u}$
\begin{equation}
\mathbf{u}=\boldsymbol{\gamma}\odot\mathbf{v}+\boldsymbol{\beta},
\end{equation}
where $\odot$ denotes element-wise multiplication. This affine modulation allows external clinical concepts to adjust internal activations without modifying the backbone. The edit vector $\mathbf{e}$, which contains clinical representations, can be obtained as
$\mathbf{e}=\mathbf{u}-\mathbf{v}\in\mathbb{R}^{d}$.

A two-layer MLP classifier $h$ then operates on $\mathbf{u}$ to predict logits $\boldsymbol{\ell}$, followed by the softmax operation to get class predicted probabilities $p$
\begin{equation}
\boldsymbol{\ell}=h(\mathbf{u}),\qquad  p=\mathrm{softmax}(\boldsymbol{\ell})\in[0,1]^2 .
\end{equation}
Given confident MONET attributes, the network is designed to learn to amplify feature channels associated with relevant concepts while attenuating others, yielding concept-consistent representations across domains.
\vspace{-.2cm}
\subsection{Stage 1: Teacher Training}
The teacher network is trained on both labeled source data and unlabeled target data. The training loss combines the supervised loss on the source domain and the consistency-based UDA losses on the target domain
\begin{equation}
\mathcal{L}_{\mathrm{teacher}}
= \mathcal{L}_{\mathrm{source}}
+ \mathcal{L}_{\mathrm{target}}
\end{equation}

\subsubsection{Source Supervision}
To address class imbalance and emphasize hard examples, we adopt the focal loss \cite{lin2017focal} which is a variant of cross-entropy that down-weights easy, high-confidence samples and places more weight on hard ones. 

\begin{equation}
\mathcal{L}_{\mathrm{sup}}
= -\frac{1}{|\mathcal{B}|}\sum_{i\in\mathcal{B}}
\alpha_{y^{(i)}}\,
\big(1-p_{y^{(i)}}\big)^{\gamma_f}\,
\log p_{y^{(i)}}
\end{equation}
where $\gamma_f=1.5$ and class weight $\alpha=0.9$, interpreted as $\alpha_{0}=0.1$ and $\alpha_{1}=0.9$ for classes $\{0,1\}$. $\mathcal{B}$ denotes the mini-batch of labeled samples used in one training step and the loss is averaged across all samples in the current batch.

During the training process, the FiLM edit vector $\mathbf{e}$ is further regularized to maintain stable and interpretable feature edits. First, an orthogonality constraint $\mathcal{L}_{\perp}$ is used to penalize edits that align with the classifier’s decision directions, preventing FiLM from trivially amplifying class-discriminative components rather than encoding clinical cues.
\begin{equation}
\mathcal{L}_{\perp} = \text{MSE}\left((\mathbf{e}\mathbf{W}_{\text{cls}}^{\top})\mathbf{W}_{\text{cls}}, \mathbf{0}\right)
\end{equation}
where $\mathbf{W}_{\mathrm{cls}}$ denotes the weight matrix of the final linear classifier $h$ and $\mathrm{MSE}$ is mean squared error. 

Additionally, a soft$-$norm constraint $\mathcal{L}_{\mathrm{norm}}$ is used to limit the overall edit magnitude, ensuring FiLM performs small, controlled adjustments to the feature representation rather than large distortions.
\begin{equation}
\mathcal{L}_{\mathrm{norm}}
\;=\;
\max\!\big(0,\ \|\mathbf{e}\|_2 - R_{\max}\big),
\quad R_{\max}=2.0
\end{equation}
Together, these regularizers encourage FiLM to make clinically meaningful refinements without altering the classifier boundary directly. 

The supervised training loss for the source domain is then
\begin{equation}
\mathcal{L}_{\mathrm{source}}
= \mathcal{L}_{\mathrm{sup}}
+ \lambda_{\perp}\,\mathcal{L}_{\perp}
+ \lambda_{\mathrm{norm}}\,\mathcal{L}_{\mathrm{norm}}
\end{equation}
where $\lambda_{\perp}=\lambda_{\mathrm{norm}}=0.01$.
%******************
\vspace{-8pt
}
\paragraph{EMA teacher} is a running average of the online model \cite{tarvainen2017mean}. It acts as a stable target during training by smoothing short-term noise from recent updates. This improves consistency in pseudo-labels and makes training more stable.

We maintain a teacher network as an exponential moving average (EMA) of the online parameters $\boldsymbol{\theta}$. At the start of training we set $\bar{\boldsymbol{\theta}}_0\!\leftarrow\!\boldsymbol{\theta}_0$, and after every optimizer update we perform
\[
\bar{\boldsymbol{\theta}} \leftarrow \alpha_{\mathrm{ema}}\bar{\boldsymbol{\theta}}
+ (1{-}\alpha_{\mathrm{ema}})\boldsymbol{\theta},\quad
\alpha_{\mathrm{ema}}=0.999,
\]
with stop-gradient on the teacher and no optimizer applied to $\bar{\boldsymbol{\theta}}$. $\alpha_{\mathrm{ema}}$ is the decay rate which defines how quickly the teacher model follows the online model. Here we set $\alpha_{\mathrm{ema}}$ to 0.999 to make sure that the teacher updates slowly, providing temporally smoothed predictions that suppress short-term training noise while still tracking long-term improvements of the online model. 
All model parameters (backbone, FiLM, head) are included in the EMA.

\subsubsection{Target Consistency Learning}
To stabilize adaptation on unlabeled target data, we adopt a two-view consistency strategy between the EMA teacher and the online MONET-guided model. For each target image, two augmentations are generated: a weak view preserving global structure and a strong view with larger appearance perturbations \cite{sohn2020fixmatch}. As both views share the same MONET-based context, we encourage agreement between weak/strong views of the same target image and stabilize the edited representation by matching both the logits and the edited features. 

The EMA teacher processes the weak view to produce a stable pseudo-label, while the online model predicts from the strong view. With temperature sharpening $\tau=0.6$, logits are converted to soft prediction $p=\mathrm{softmax}(\boldsymbol{\ell}/{\tau})$. Consistency is enforced by minimizing the divergence between the EMA teacher’s weak-view prediction $\bar{p}_w$ and the model's strong-view prediction $p_s$.
\begin{equation}
\mathcal{L}_{\mathrm{KL}}
= 0.5 \left(
\mathrm{KL}\!\big(p_s \,\|\, \bar{p}_w\big)
+
\mathrm{KL}\!\big(\bar{p}_w \,\|\, p_s\big)
\right)
\end{equation}
Feature- and edit-level consistencies are also maintained using MSE losses
\begin{equation}
\mathcal{L}_{\mathrm{feat}}=\mathrm{MSE}(\mathbf{u}_s,\mathbf{\bar{u}}_w),\qquad
\mathcal{L}_{\mathrm{edit}}=\mathrm{MSE}(\mathbf{e}_s,\mathbf{\bar{e}}_w).
\end{equation}

To reduce noise from unreliable pseudo-labels, only target samples with sufficiently confident predictions are used in the loss. In this paper, we use a dynamic confidence threshold $t(\mathrm{epoch})$ to control sample selection. Early in training, $t(0)=0.95$ is set to restrict learning to high-confidence examples and the threshold gradually decays to 0.70 as the teacher becomes more stable. This strategy prevents error propagation at the start and increases data utilization over time.

Then the total target loss combines the masked, two-view consistency terms
\begin{equation}
\mathcal{L}_{\mathrm{target}}
= w_{\mathrm{kl}}\,\mathcal{L}_{\mathrm{KL}}
+ w_{\mathrm{feat}}\,\mathcal{L}_{\mathrm{feat}}
+ w_{\mathrm{edit}}\,\mathcal{L}_{\mathrm{edit}}
\end{equation}
with $w_{\mathrm{kl}}=0.6$, $w_{\mathrm{feat}}=0.1$, and $w_{\mathrm{edit}}=0.1$. These losses encourage the online model to match the EMA teacher’s stable weak-view predictions under strong perturbations, yielding robust, domain-invariant representations. 
\subsection{Stage~2: Image-Only Student Distillation}\label{sec:student} After training the concept-guided EMA teacher, an image-only student network is trained to reproduce the teacher's behavior without MONET inputs. We denote the teacher by $T$ and the student by $S$. The teacher parameters $\bar{\boldsymbol{\theta}}$ are frozen while the student parameters ${\boldsymbol{\theta}}_S$ are trained independently to imitate the teacher using only image input.
Given an image $x$, the EMA teacher produces the concept-guided feature $\mathbf{u}_T$ and corresponding logits $\boldsymbol{\ell}_T$. The student network shares the same backbone and classifier but replaces the FiLM module with a lightweight residual edit head $\boldsymbol{\psi}_{\mathrm{edit}}$ that predicts its edited feature directly from the image
\begin{equation}
\mathbf{v}_S=f_{\theta_S}(\mathbf{x}),\quad
\mathbf{u}_S=\mathbf{v}_S+\boldsymbol{\psi}_{\mathrm{edit}}(\mathbf{v}_S).
\end{equation}
Then the student classifier outputs logits $\boldsymbol{\ell}_S=h(\mathbf{u}_S)$ and the teacher and student logits are converted into soft probability distributions $p_S$, $p_T$ using a temperature-scaled softmax where the temperature $\tau$ set to 2.0 to smooth the distributions to reveal relative class confidence \cite{hinton2015distilling}. The student loss is 
\begin{equation}
\mathcal{L}_{\mathrm{distill}}
= \mathrm{KL}\!\big(p_S\|p_T\big)\,\tau^2
+ \lambda_{\mathrm{feat}}\,\mathrm{MSE}(\mathbf{u}_S,\mathbf{u}_T)
\end{equation}
with $\lambda_{\mathrm{feat}}=0.1$. The KL-divergence loss transfers the teacher’s class-probability structure, and the feature-matching term encourages the student to reproduce the teacher’s concept-aware representation.
Through this process, the student inherits the teacher’s clinically informed reasoning yet remains independent of MONET at inference, enabling efficient and domain-robust image-only prediction.
\subsection{Stage~3: Image-Only Inference and Evaluation}\label{sec:stage3}
At inference time, the image-only student model is used. The predicted melanoma probability $p_1$ determines the final label, where melanoma is predicted if $p_1 \geq 0.5$. Neither MONET concept inputs nor EMA teacher updates are required during inference, ensuring a lightweight and fully image-based inference.
\end{tightmath}
}

\vspace{-5pt}
\section{Results and Discussion}
\begin{figure}[H]
\captionsetup{aboveskip=2pt, belowskip=-5pt}
    \centering    \includegraphics[width=0.45\textwidth,height=0.15\textheight]{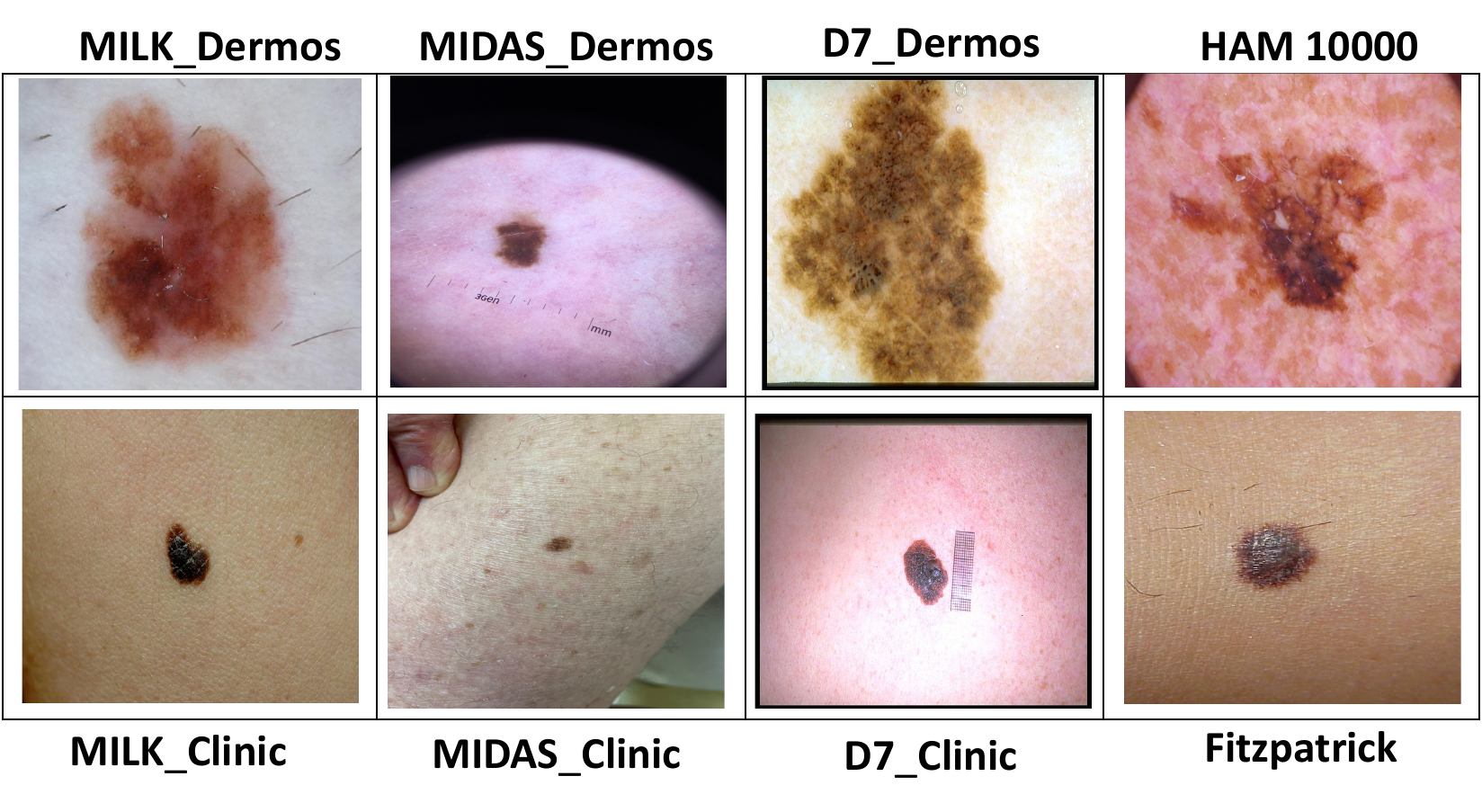}
    \caption{Eight dataset examples across dermoscopic and clinical capture. Dermoscopic images are close and uniform with fine texture. Clinical images include wider context, varied lighting, and common artefacts such as hair, rulers, and shadows. These differences create a clear domain gap for the same lesion classes.}
    \label{dataset}
\end{figure}

\vspace{-5pt}
\paragraph{Datasets:}
We use eight public skin lesion datasets encompassing two capture styles, Dermoscopic and Clinical, with examples shown in Figure \ref{dataset} and a detailed summary in Table \ref{tab:datasets}.
During training, image-level concept probabilities derived from the clinical decision-support tool (MONET) are incorporated as supervision. These concept probabilities are not used during inference. 

\paragraph{Evaluation Protocol:}
We train on a single source–target pair, \textbf{MILK Dermoscopic} $\rightarrow$ \textbf{MILK Clinical}. 
After training the teacher and student model separately, the resulting image-only student model is frozen and evaluated directly on the test partitions of six unseen external datasets. 
This is a strict out-of-domain evaluation. Results are reported using AUROC and melanoma recall (sensitivity), with high recall being critical to ensure potential melanomas are not missed during screening.

\paragraph{Reproducibility:} 
All experiments are run end-to-end with five random seeds. Models train for up to 50 epochs with early stopping on validation AUROC using development data only, ensuring strict separation from the test sets. Inference uses images only. Both teacher and student models use batch size 32, an EfficientNet-B2 backbone \cite{tan2019efficientnet}, AdamW (lr $3\times10^{-4}$, weight decay $10^{-4}$), and cosine annealing to $10^{-6}$. Weak and strong augmentations (crop, flip, color jitter, Gaussian blur) are applied uniformly across methods. Additional metrics, plots, ablations, and implementation details are provided in the supplementary material. Experiments are performed on an Apple Mac~Studio (M3~Ultra, 28-core CPU, 60-core GPU, 32-core Neural Engine) with 96\,GB of unified memory. 

\begin{table}[H]
\centering
\captionsetup{aboveskip=2pt, belowskip=-5pt}
\caption{Dataset summary across eight datasets. Bold\textsuperscript{*} datasets denote unseen evaluation sets that are never used for training or adaptation. \textbf{Mel} counts all melanoma subtypes, and \textbf{Other} includes all remaining diagnostic categories. The training dataset is split 70:15:15 into training, validation, and test partitions. Each unseen dataset uses all available images exclusively for test-time inference with the image-only model.}
\label{tab:datasets}
\resizebox{\linewidth}{!}{%
\begin{tabular}{l l r r}
\hline
Dataset & Domain & Mel & Other \\
\hline
MILK10K (Md / Mc)\cite{philipp2025milk10k}      & Dermoscopic and Clinical & 452 & 4,792 \\
\textbf{Derm7pt} \textsuperscript{*}  (D7d / D7c)\cite{kawahara2019derm7pt}   & Dermoscopic and Clinical & 254 &   762 \\
\textbf{Fitzpatrick}  (Fitz) \textsuperscript{*}\cite{fitzpatrick1988skintype}& Clinical                 & 344 & 11,259 \\
\textbf{HAM10000} \textsuperscript{*}(HAM)\cite{tschandl2018ham10000}       & Dermoscopic              & 780 &  6,232 \\
\textbf{MIDAS} \textsuperscript{*} (MID\_d)\cite{chiou2025multimodal}        & Dermoscopic              &  96 &    952 \\
\textbf{MIDAS} \textsuperscript{*} (MID\_c)\cite{chiou2025multimodal}        & Clinical                 & 186 &  1,915 \\
\hline
\end{tabular}%
}
\end{table}

\paragraph{Baselines:}
We compare our approach to `source-only' training and a wide range of domain-adaptation and adaptation-free baselines, including 
DANN~\cite{ganin2016domain}, Deep~CORAL~\cite{sun2016deep}, MMD~\cite{gretton2012kernel}, Mean~Teacher~\cite{tarvainen2017mean}, 
IT-RUDA~\cite{rashidi2025ruda}, SHOT++~\cite{liang2021source}, SFDA~\cite{liu2021source}, 
CPD~\cite{zhou2024source}, DINE~\cite{liang2022dine}, CoTTA~\cite{wang2022continual}, 
WoC~\cite{lee2023towards}, TENT~\cite{wang2021tent}, DAMP~\cite{du2024domain}, and DALUPI~\cite{Breitholtz2024DALUPI}. 
Training data, augmentation policies, and schedules are matched where applicable. 
Unless otherwise stated, all reported results correspond to the final image-only student model.
\vspace{-0.9em}

~\input{joint_auroc_table}

\paragraph{Comparison with other Methods:}
We evaluate our distilled image-only student (CoFiDA-M) against the source only baseline and competing UDA methods (Tables~\ref{tab:AUROC}). As shown in Table~\ref{tab:AUROC}(B), our method achieves the highest average AUROC on the clinical target domain at 67.50\%. This represents a +9.11\% absolute improvement over the source only baseline (58.39\%) and significantly exceeds other strategies, including the strongest test-time method, TENT (62.32\%), and the privileged information baseline, DALUPI (54.54\%).

For the critical screening metric of melanoma recall Table~\ref{tab:recall}, the improvement is more pronounced. On the clinical target domain Table~\ref{tab:recall}(B), CoFiDA-M achieves an average recall of 77.89\%, a +22.12\% absolute gain over the source only baseline (55.77\%). Notably, while the standard UDA method Mean Teacher achieves a similar high recall (77.67\%), its clinical AUROC (58.67\%) remains near the baseline. In contrast, our method yields substantial improvements on both target-domain metrics. Importantly, the gain does not arise from a trade-off: our method also achieves the highest average recall on the source dermoscopic domain Table~\ref{tab:AUROC}(A) reaching 84.92\% compared to 63.55\% for the source-only baseline. These results indicate that our model learns a more robust, domain-invariant representation of the minority class, enabling safer and more reliable cross-domain generalization.
~

\input{abla}

\begin{figure}[h]
\captionsetup{aboveskip=0pt, belowskip=-25pt}
    \centering
\includegraphics[width=.5\textwidth]{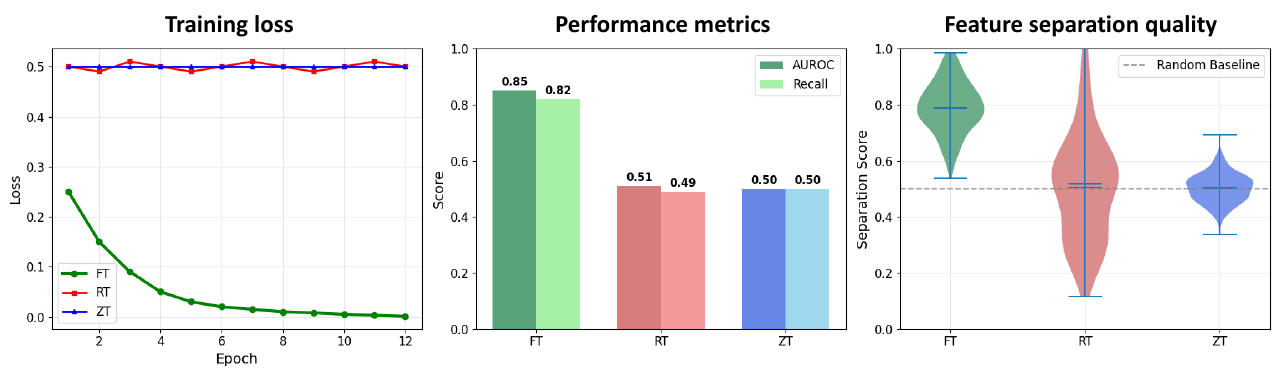}
    \caption{Ablation on Teacher Knowledge on MILK (Clinical). The image-only student is distilled from three fixed teachers: a fully trained Frozen Teacher (FT), a Random-weight Teacher (RT), and a Zero-weight Teacher (ZT).}
    \label{Sanity}
\end{figure}

\begin{figure}[h]
\captionsetup{aboveskip=2pt, belowskip=-20pt}
\centering  \includegraphics[width=.48\textwidth]{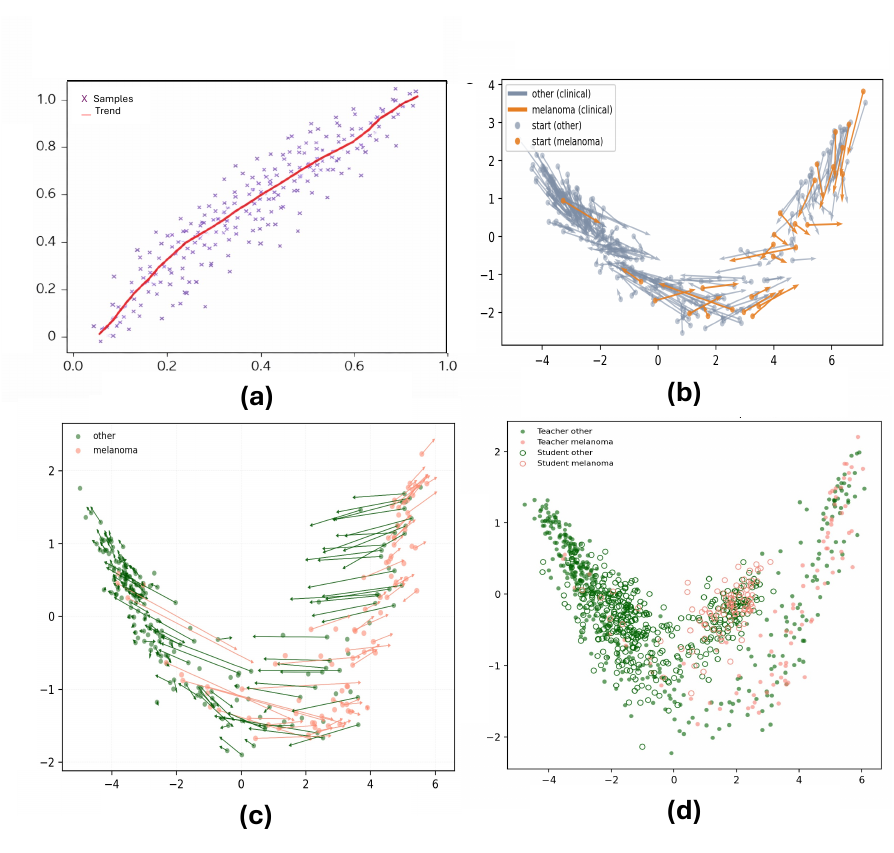}
\caption{Qualitative Validation of the CoFiDA-M Framework. (a) Student's implicit concept learning via edit magnitude correlation. (b) Teacher's feature-editing mechanism transforming feature space. (c) Causal proof of concept-guided feature steering. (d) Successful distillation via Student-Teacher feature alignment.}
    \label{quality}
\end{figure}
\vspace{-8pt}

\paragraph{Ablation Study:} Table~\ref{tab:single_ablation} validates our core design choices. Section A shows our image-only student outperforms both the `Source-Only` baseline and standard UDA (without PI), confirming that distilling MONET-derived concepts provides a stronger training signal than raw unlabeled data alone. Section B analyzes the teacher's architecture. Removing the confidence-gating mechanism causes a drastic drop in melanoma recall, demonstrating the necessity of filtering noisy concept scores. Replacing FiLM-based modulation with naïve concatenation also degrades performance, confirming that explicitly editing the feature space is superior to simple feature augmentation. Finally, Section C isolates our distillation strategy. Training with only logit-based KD is insufficient. Aligning the student to the teacher's pre-edit features ($\mathbf{v}_T$) improves performance, but aligning to the post-edit, concept-adjusted features ($\mathbf{u}_T$) yields the best results. This confirms our primary hypothesis: the student must learn to reproduce the teacher's concept-adjusted representation. This finding is validated by a sanity check, Figure \ref{Sanity}, where distilling from a random-weight teacher (RT) or zero-weight teacher (ZT) fails to converge (FT $\gg$ RT $\approx$ ZT), achieving only chance-level performance. The student's gains thus arise from the teacher's learned semantic edits, not the distillation process itself. The violin plot shows that the FT-based student achieves distinctly higher and tighter feature separation. RT and ZT collapse near the random baseline, which is evidence that only the trained teacher imparts meaningful semantic structure to the student’s embeddings.

\begin{comment}
\paragraph{Statistical Significance:}
All methods are evaluated over five seeds to account for stochastic training effects. A paired Wilcoxon test on seed-wise deltas shows CoFIDA-M consistently outperforms 14 baselines (+9–19\,pp AUROC/Recall); with $n{=}5$, the test is under-powered ($p{\approx}0.12$, BH-adjusted $q{\approx}0.27$), indicating large, stable effects despite limited sample power.
\vspace{-13pt}
\end{comment}
\vspace{-.1cm}
\paragraph{Feature Space and Reliability Analysis:} Figure \ref{quality}(a) shows that the image-only student has internalized the teacher’s concept guided behavior: the MONET concept score (ground truth) correlates strongly and positively with the student’s self-generated edit magnitude \(\lVert \mathbf{u}_S - \mathbf{v}_S \rVert_2\). Figure \ref{quality}(b) visualizes the MONET-guided teacher FiLM edit \((\mathbf{v}_T \!\to\! \mathbf{u}_T)\) correcting the feature space and moving scattered samples into two well-separated clusters. Figure Figure \ref{quality}(c) provides intervention based evidence: increasing melanoma like concept input moves the FiLM-edited feature towards the melanoma cluster. Figure \ref{quality}(d) confirms successful distillation, with the image-only student (open circles) reproducing the teacher’s (filled dots) corrected, separated manifold.

\section{Conclusion}
%\vspace{-2pt}
We introduced a practical domain adaptation framework that learns from privileged, high-level concepts during training but deploys as a lightweight, image-only model. Our method uses a MONET-conditioned, FiLM-based teacher to shape the feature space using clinical semantics. We then distill not only the final logits but also the edited feature representation into a student, effectively embedding expert knowledge into its weights to enable robust, concept-free generalization at inference time. Our analysis shows that alignment with the edited semantic space is critical for transferring the teacher’s knowledge. While we demonstrated this with MONET in dermatology, the core privileged-information framework is general. The main limitation lies in its dependency on the quality of training-time concepts, which future work could address by learning concepts end-to-end or extending the framework to other specialized domains.

\bibliographystyle{ieeenat_fullname}
\bibliography{main}

\end{document}

%% file: joint_auroc_table.tex
% PREAMBLE: \usepackage{caption}  % needed for \captionof

\begingroup
\sisetup{
  separate-uncertainty = true,
  table-number-alignment = center,
  detect-weight = true,
  detect-family = true
}
\setlength{\tabcolsep}{4pt}

\begin{table*}[!t]
\captionsetup{aboveskip=2pt, belowskip=-10pt}
\centering
% ===================== (1) AUROC TABLE =====================
\begin{minipage}{\linewidth}
\centering
\scriptsize
\renewcommand{\arraystretch}{1.05}
\captionof{table}{AUROC (\%) mean $\pm$ std over \textbf{5} seeds. Left: \textbf{Dermoscopic} (A), macro-average across Dermoscopic and Right \textbf{Clinical} (B), macro-average across Clinical. MIDd, MIDc, D7d, D7c, HAM and Fitz are unseen test datasets.}
\label{tab:AUROC}
\resizebox{\linewidth}{!}{%
\begin{tabular}{
  l
  S[table-format=2.2(2)]
  S[table-format=2.2(2)]
  S[table-format=2.2(2)]
  S[table-format=2.2(2)]
  S[table-format=2.2]
  !{\vrule width 1.4pt}
  S[table-format=2.2(2)]
  S[table-format=2.2(2)]
  S[table-format=2.2(2)]
  S[table-format=2.2(2)]
  S[table-format=2.2]
}
\toprule
& \multicolumn{5}{c}{\textbf{(A) Dermoscopic}} & \multicolumn{5}{c}{\textbf{(B) Clinical}} \\
\cmidrule(lr){2-6} \cmidrule(lr){7-11}
\textbf{Setting / Method} & {\textbf{Md}} & {\textbf{MIDd}} \textsuperscript{*} & {\textbf{D7d}} \textsuperscript{*} & {\textbf{HAM}} \textsuperscript{*} & {\textbf{Avg}} & {\textbf{Mc}} & {\textbf{MIDc}}\textsuperscript{*} & {\textbf{D7c}}\textsuperscript{*} &{\textbf{Fitz}}\textsuperscript{*} & {\textbf{Avg}} \\
\midrule
\multicolumn{11}{l}{\textbf{Classical / Standard}} \\
DANN \cite{ganin2016domain}
& \num{78.37 \pm 0.05} & \num{72.29 \pm 0.08} & \num{69.40 \pm 0.07} & \num{72.09 \pm 0.09} & 73.04
& \num{76.81 \pm 0.05} & \num{56.43 \pm 0.07} & \num{43.84 \pm 0.06} & \num{55.23 \pm 0.09} & 58.08 \\
DeepCoral \cite{sun2016deep}
& \num{79.20 \pm 0.05} & \num{69.55 \pm 0.07} & \num{52.39 \pm 0.04} & \num{76.87 \pm 0.06} & 69.50
& \num{77.30 \pm 0.06} & \num{50.41 \pm 0.12} & \num{49.37 \pm 0.07} & \num{48.76 \pm 0.04} & 56.46 \\
MMD \cite{gretton2012kernel}
& \num{78.23 \pm 0.09} & \num{75.83 \pm 0.10} & \num{62.15 \pm 0.16} & \num{80.10 \pm 0.17} & 74.08
& \num{79.58 \pm 0.04} & \num{52.25 \pm 0.05} & \num{46.74 \pm 0.12} & \num{56.10 \pm 0.09} & 58.67 \\
MeanTeacher \cite{tarvainen2017mean}
& \num{62.06 \pm 0.26} & \num{45.34 \pm 0.13} & \num{51.22 \pm 0.10} & \num{65.27 \pm 0.12} & 55.97
& \num{53.08 \pm 0.20} & \num{48.81 \pm 0.05} & \num{47.46 \pm 0.09} & \num{46.30 \pm 0.08} & 48.91 \\
IT\texttt{-}RUDA \cite{rashidi2025ruda}
& \num{86.27 \pm 0.07} & \num{67.16 \pm 0.02} & \num{67.19 \pm 0.05} & \num{82.79 \pm 0.06} & 75.85
& \num{63.34 \pm 0.06} & \num{51.68 \pm 0.05} & \num{57.39 \pm 0.08} & \num{53.65 \pm 0.01} & 56.52 \\
\midrule
\multicolumn{11}{l}{\textbf{Modern Source Free}} \\
SHOT++ \cite{liang2021source}
& \num{65.17 \pm 0.14} & \num{52.57 \pm 0.09} & \num{62.19 \pm 0.19} & \num{61.45 \pm 0.15} & 60.35
& \num{55.88 \pm 0.03} & \num{57.39 \pm 0.02} & \num{51.10 \pm 0.11} & \num{39.01 \pm 0.09} & 50.84 \\
DINE \cite{liang2022dine}
& \num{70.32 \pm 0.13} & \num{71.80 \pm 0.12} & \num{65.39 \pm 0.12} & \num{79.81 \pm 0.10} & 71.83
& \num{66.54 \pm 0.15} & \num{52.93 \pm 0.08} & \num{41.22 \pm 0.07} & \num{56.94 \pm 0.07} & 54.41 \\
SFDA \cite{liu2021source}
& \num{58.49 \pm 0.11} & \num{55.76 \pm 0.16} & \num{55.95 \pm 0.03} & \num{54.12 \pm 0.19} & 56.08
& \num{44.91 \pm 0.12} & \num{47.27 \pm 0.05} & \num{45.39 \pm 0.07} & \num{51.42 \pm 0.10} & 47.25 \\
CPD \cite{zhou2024source}
& \num{59.87 \pm 0.09} & \num{59.87 \pm 0.06} & \num{57.05 \pm 0.02} & \num{69.24 \pm 0.07} & 61.51
& \num{46.14 \pm 0.05} & \num{51.40 \pm 0.06} & \num{52.83 \pm 0.05} & \num{42.54 \pm 0.07} & 48.23 \\
\midrule
\multicolumn{11}{l}{\textbf{Modern Test Time}} \\
CoTTA \cite{wang2022continual}
& \num{62.43 \pm 0.13} & \num{53.60 \pm 0.22} & \num{55.79 \pm 0.01} & \num{54.61 \pm 0.07} & 56.61
& \num{51.99 \pm 0.14} & \num{50.55 \pm 0.15} & \num{53.71 \pm 0.06} & \num{52.75 \pm 0.09} & 52.25 \\
TENT \cite{wang2021tent}
& \num{80.21 \pm 0.16} & \num{65.30 \pm 0.04} & \num{70.08 \pm 0.09} & \num{63.21 \pm 0.16} & 69.70
& \num{81.76 \pm 0.11} & \num{55.09 \pm 0.17} & \num{55.78 \pm 0.15} & \num{56.63 \pm 0.07} & 62.32 \\
WoC \cite{lee2023towards}
& \num{66.29 \pm 0.18} & \num{58.19 \pm 0.10} & \num{61.05 \pm 0.20} & \num{54.71 \pm 0.14} & 60.06
& \num{49.76 \pm 0.13} & \num{43.57 \pm 0.16} & \num{60.73 \pm 0.12} & \num{41.87 \pm 0.02} & 48.98 \\
\midrule
\multicolumn{11}{l}{\textbf{Multimodal}} \\
DAMP \cite{du2024domain}
& \num{87.32 \pm 0.09} & \num{65.57 \pm 0.02} & \num{69.02 \pm 0.01} & \num{80.08 \pm 0.01} & 75.50
& \num{63.59 \pm 0.01} & \num{43.52 \pm 0.03} & \num{61.14 \pm 0.04} & \num{53.68 \pm 0.05} & 55.48 \\
DALUPI \cite{Breitholtz2024DALUPI}
& \num{87.38 \pm 0.01} & \num{62.69 \pm 0.03} & \num{70.26 \pm 0.07} & \num{83.97 \pm 0.01} & 76.07
& \num{58.18 \pm 0.02} & \num{47.71 \pm 0.03} & \num{62.50 \pm 0.08} & \num{49.78 \pm 0.02} & 54.54 \\
\midrule
Source Only
& \num{84.67 \pm 0.06} & \num{65.94 \pm 0.03} & \num{54.76 \pm 0.05} & \num{74.48 \pm 0.01} & 69.96
& \num{75.02 \pm 0.04} & \num{51.51 \pm 0.02} & \num{52.85 \pm 0.08} & \num{54.20 \pm 0.01} & 58.39 \\
\textbf{Ours (CoFIDA-M)}
& \num{89.70 \pm 0.09} & \num{76.15 \pm 0.07} & \num{61.69 \pm 0.04} & \num{78.45 \pm 0.05} & \textbf{76.50}
& \num{83.94 \pm 0.01} & \num{69.67 \pm 0.04} & \num{50.48 \pm 0.07} & \num{65.91 \pm 0.09} & \textbf{67.50} \\
\bottomrule
\end{tabular}%
}
\end{minipage}

\vspace{10pt}

% ===================== (2) RECALL TABLE =====================
\begin{minipage}{\linewidth}
\captionsetup{aboveskip=2pt, belowskip=-8pt}
\centering
\scriptsize
\renewcommand{\arraystretch}{1.05}
\captionof{table}{Recall for melanoma (\%) mean $\pm$ std over \textbf{5} seeds. Left: \textbf{Dermoscopic} (A), macro-average across Dermoscopic and Right \textbf{Clinical} (B), macro-average across Clinical. MIDd, MIDc, D7d, D7c, HAM and Fitz are unseen test datasets.}
\label{tab:recall}
\resizebox{\linewidth}{!}{%
\begin{tabular}{
  l
  S[table-format=2.2(2)]
  S[table-format=2.2(2)]
  S[table-format=2.2(2)]
  S[table-format=2.2(2)]
  S[table-format=2.2]
  !{\vrule width 1.4pt}
  S[table-format=2.2(2)]
  S[table-format=2.2(2)]
  S[table-format=2.2(2)]
  S[table-format=2.2(2)]
  S[table-format=2.2]
}
\toprule
& \multicolumn{5}{c}{\textbf{(A) Dermoscopic}} & \multicolumn{5}{c}{\textbf{(B) Clinical}} \\
\cmidrule(lr){2-6} \cmidrule(lr){7-11}
\textbf{Setting / Method} & {\textbf{Md}} & {\textbf{MIDd}}\textsuperscript{*} & {\textbf{D7d}}\textsuperscript{*} & {\textbf{HAM}}\textsuperscript{*} & {\textbf{Avg}} & {\textbf{Mc}} & {\textbf{MIDc}}\textsuperscript{*} & {\textbf{D7c}}\textsuperscript{*} & {\textbf{Fitz}}\textsuperscript{*} & {\textbf{Avg}} \\
\midrule
\multicolumn{11}{l}{\textbf{Classical / Standard}} \\
DANN\cite{ganin2016domain}
& \num{40.29 \pm 0.17} & \num{70.89 \pm 0.02} & \num{38.30 \pm 0.09} & \num{62.87 \pm 0.03} & 53.09
& \num{21.59 \pm 0.05} & \num{49.86 \pm 0.12} & \num{20.45 \pm 0.06} & \num{03.47 \pm 0.06} & 23.84 \\
DeepCoral\cite{sun2016deep}
& \num{79.20 \pm 0.24} & \num{69.55 \pm 0.19} & \num{52.39 \pm 0.31} & \num{76.87 \pm 0.22} & 69.50
& \num{31.82 \pm 0.02} & \num{10.97 \pm 0.04} & \num{50.00 \pm 0.02} & \num{14.86 \pm 0.05} & 26.91 \\
MMD\cite{gretton2012kernel}
& \num{43.69 \pm 0.24} & \num{55.70 \pm 0.01} & \num{51.06 \pm 0.02} & \num{86.46 \pm 0.07} & 59.23
& \num{30.68 \pm 0.07} & \num{05.81 \pm 0.01} & \num{27.27 \pm 0.02} & \num{09.46 \pm 0.10} & 18.31 \\
MeanTeacher\cite{tarvainen2017mean}
& \num{74.91 \pm 0.32} & \num{70.89 \pm 0.04} & \num{65.96 \pm 0.08} & \num{44.31 \pm 0.07} & 64.02
& \num{77.27 \pm 0.01} & \num{88.71 \pm 0.04} & \num{70.45 \pm 0.07} & \num{74.23 \pm 0.05} & 77.67 \\
IT-RUDA\cite{rashidi2025ruda}
& \num{49.28 \pm 0.02} & \num{12.66 \pm 0.01} & \num{12.62 \pm 0.09} & \num{47.90 \pm 0.01} & 30.62
& \num{9.77 \pm 0.06} & \num{7.56 \pm 0.07} & \num{12.34 \pm 0.09} & \num{22 \pm 0.10} & 12.92 \\
\midrule
\multicolumn{11}{l}{\textbf{Modern Source Free}} \\
SHOT++\cite{liang2021source}
& \num{72.61 \pm 0.09} & \num{64.56 \pm 0.01} & \num{39.26 \pm 0.19} & \num{68.26 \pm 0.008} & 61.17
& \num{65.91 \pm 0.01} & \num{57.42 \pm 0.03} & \num{52.94 \pm 0.01} & \num{54.05 \pm 0.01} & 57.58 \\
DINE\cite{liang2022dine}
& \num{02.27 \pm 0.01} & \num{08.86 \pm 0.02} & \num{09.80 \pm 0.05} & \num{38.92 \pm 0.03} & 14.96
& \num{02.27 \pm 0.01} & \num{03.58 \pm 0.03} & \num{01.96 \pm 0.01} & \num{01.35 \pm 0.02} & 2.29 \\
SFDA\cite{liu2021source}
& \num{24.53 \pm 0.21} & \num{22.09 \pm 0.13} & \num{20.10 \pm 0.09} & \num{19.77 \pm 0.15} & 21.62
& \num{06.31 \pm 0.06} & \num{10.67 \pm 0.12} & \num{15.91 \pm 0.01} & \num{09.34 \pm 0.10} & 10.56 \\
CPD\cite{zhou2024source}
& \num{57.95 \pm 0.01} & \num{35.44 \pm 0.06} & \num{70.86 \pm 0.06} & \num{62.28 \pm 0.09} & 56.63
& \num{55.68 \pm 0.04} & \num{68.23 \pm 0.06} & \num{71.52 \pm 0.02} & \num{47.52 \pm 0.19} & 60.74 \\
\midrule
\multicolumn{11}{l}{\textbf{Modern Test Time}} \\
CoTTA\cite{wang2022continual}
& \num{68.29 \pm 0.07} & \num{59.49 \pm 0.03} & \num{58.82 \pm 0.04} & \num{69.58 \pm 0.09} & 64.05
& \num{56.82 \pm 0.01} & \num{63.23 \pm 0.03} & \num{52.94 \pm 0.08} & \num{59.46 \pm 0.05} & 58.11 \\
TENT\cite{wang2021tent}
& \num{76.11 \pm 0.10} & \num{67.24 \pm 0.09} & \num{50.22 \pm 0.27} & \num{69.37 \pm 0.17} & 65.74
& \num{75.00 \pm 0.06} & \num{61.10 \pm 0.29} & \num{48.30 \pm 0.20} & \num{38.22 \pm 0.11} & 55.65 \\
WoC\cite{lee2023towards}
& \num{52.45 \pm 0.07} & \num{34.18 \pm 0.01} & \num{68.09 \pm 0.05} & \num{38.92 \pm 0.09} & 48.41
& \num{37.50 \pm 0.02} & \num{09.03 \pm 0.01} & \num{61.36 \pm 0.07} & \num{05.41 \pm 0.05} & 28.33 \\
\midrule
\multicolumn{11}{l}{\textbf{Multimodal}} \\
DAMP\cite{du2024domain}
& \num{35.23 \pm 0.07} & \num{15.19 \pm 0.03} & \num{13.25 \pm 0.02} & \num{35.93 \pm 0.06} & 24.90
& \num{1.14 \pm 0.05} & \num{09.32 \pm 0.03} & \num{07.28 \pm 0.05} & \num{5.57 \pm 0.06} & 5.83 \\
DALUPI\cite{Breitholtz2024DALUPI}
& \num{53.41 \pm 0.07} & \num{29.11 \pm 0.05} & \num{25.83 \pm 0.09} & \num{52.10 \pm 0.10} & 40.11
& \num{04.55 \pm 0.06} & \num{07.74 \pm 0.05} & \num{12.58 \pm 0.08} & \num{13.99 \pm 0.01} & 9.72 \\
\midrule
Source Only
& \num{72.37 \pm 0.04} & \num{62.70 \pm 0.22} & \num{58.64 \pm 0.02} & \num{60.48 \pm 0.12} & 63.55
& \num{65.91 \pm 0.03} & \num{65.94 \pm 0.03} & \num{55.78 \pm 0.14} & \num{35.45 \pm 0.04} & 55.77 \\
\textbf{Ours (CoFIDA-M)}
& \num{81.56 \pm 0.11} & \num{75.95 \pm 0.01} & \num{89.36 \pm 0.02} & \num{92.81 \pm 0.07} & \textbf{84.92}
& \num{86.36 \pm 0.02} & \num{74.84 \pm 0.01} & \num{90.91 \pm 0.02} & \num{59.46 \pm 0.03} & \textbf{77.89} \\
\bottomrule
\end{tabular}%
}
\end{minipage}
\end{table*}
\endgroup
%\vspace{-15pt}

%% file: abla.tex
%**************************************

\begin{table*}[t]
    \captionsetup{aboveskip=0pt}
    \centering
    \caption{Ablation study. Each metric reflects the average percentage score across the four Dermoscopic (d) and four Clinical (c) datasets. Section A compares image-only frameworks, Section B varies the teacher architecture, and Section C examines student distillation using the same frozen teacher.}
    \label{tab:single_ablation}
    \small
    
    % --- Squish controls (width + row gap) ---
    \setlength{\tabcolsep}{3.5pt}           % tighter column padding (default ~6pt)
    \renewcommand{\arraystretch}{0.92}       % tighter row height
    \setlength{\aboverulesep}{0pt}           % tighten booktabs spacing
    \setlength{\belowrulesep}{0pt}
    \setlength{\cmidrulekern}{0.2em}         % thinner midrule gaps
    
    % Fixed widths for text columns (adjust if needed)
    \newcommand{\secw}{0.6cm}
    \newcommand{\famw}{2.6cm}
    \newcommand{\varw}{4.5cm}
    
    \begin{tabular}{
        @{}                                     % kill outer left padding
        >{\centering\arraybackslash}p{\secw}    % Sec.
        >{\raggedright\arraybackslash}p{\famw}  % Family
        >{\raggedright\arraybackslash}p{\varw}  % Variant
        S[table-format=2.2]                     % AUROC_Derm (fits 2.2 safely)
        S[table-format=2.2]                     % AUROC_Clin
        S[table-format=2.2]                     % Recall_Derm
        S[table-format=2.2]                     % Recall_Clin
        @{}                                     % kill outer right padding
    }
    \toprule
    \multicolumn{1}{c}{\textbf{Sec.}} &
    \multicolumn{1}{c}{\textbf{Family}} &
    \multicolumn{1}{c}{\textbf{Variant}} &
    \multicolumn{1}{c}{{AUROC$_{\text{d}}$ $\uparrow$}} &
    \multicolumn{1}{c}{{AUROC$_{\text{c}}$}$\uparrow$} &
    \multicolumn{1}{c}{{Recall$_{\text{d}}$}$\uparrow$} &
    \multicolumn{1}{c}{{Recall$_{\text{c}}$}$\uparrow$} \\
    \midrule
    \multirow{3}{*}{A} & \multirow{3}{*}{Baselines / Student}
      & Source-Only                               & 69.96 & 58.39 & 63.55 & 55.77 \\
     &  & Standard UDA (no MONET)                  & 76.07 & 62.32 & 69.50 & 77.67 \\
     &  & \textbf{Ours: Student (image-only)}      & \bfseries 76.50 & \bfseries 67.50 & \bfseries 84.92 & \bfseries 77.89 \\
    \midrule
    \multirow{3}{*}{B} & \multirow{3}{*}{Teacher design}
      & Full Teacher (Gating + FiLM)               & \bfseries 85.35 & \bfseries 83.81 & \bfseries 87.81 & \bfseries 86.36 \\
     &  & w/o Confidence-Gating                     & 78.32 & 75.24 & 47.70 & 43.20 \\
     &  & w/o FiLM (Concat)                         & 84.18 & 81.51 & 80.68 & 73.86 \\
    \midrule
    \multirow{3}{*}{C} & \multirow{3}{*}{\parbox[t]{\famw}{\raggedright Knowledge distillation (fixed Teacher)}}
      & Logit KD Only                               & 63.80 & 64.10 & 72.49 & 64.53 \\
     &  & Logit KD + Align $\mathbf{v}_T$ (pre-edit)         & 70.68 & 66.45 & 79.12 & 75.74 \\
     &  & \textbf{Logit KD + Align $\mathbf{u}_T$ (post-edit)} & \bfseries 76.50 & \bfseries 67.50 & \bfseries 84.92 & \bfseries 77.89 \\
    \bottomrule
    \end{tabular}
\end{table*}